\documentclass[11pt]{article}
\usepackage{naaclhlt2016}
\usepackage{times}
\usepackage{url}
\usepackage{latexsym}
\usepackage{xspace}
\usepackage{booktabs}
\usepackage{color}
\usepackage{graphicx}
\usepackage{tabulary}
\usepackage{enumitem}

\naaclfinalcopy

\usepackage{amsfonts}

  {\begin{itemize}[topsep=0pt, partopsep=0pt] %
    \setlength{\itemsep}{0pt}%
    \setlength{\parskip}{0pt}%
    }%
  {\end{itemize}}
  
  \usepackage{color}
  {\begin{enumerate}≈ \footnotesize%
    \setlength{\itemsep}{0pt}%
    \setlength{\parskip}{0pt}%
    }%
  {\end{enumerate}}

\usepackage{microtype}
\usepackage{multirow}
\usepackage{verbatim}
\usepackage{amsmath,amsthm,amssymb}
\usepackage{array}
\usepackage[scaled=0.86]{helvet}
\usepackage{ifthen}
\usepackage{courier}
\usepackage[linesnumbered,vlined,ruled]{algorithm2e}

\newcommand{\captionfonts}{\small}
\makeatletter  %
\long\def\@makecaption#1#2{%
  \vskip\abovecaptionskip
  \sbox\@tempboxa{{\captionfonts #1: #2}}%
  \ifdim \wd\@tempboxa >\hsize
    {\captionfonts #1: #2\par}
  \else
    \hbox to\hsize{\hfil\box\@tempboxa\hfil}%
  \fi
  \vskip\belowcaptionskip}
\makeatother   %

\belowdisplayskip \abovedisplayskip

\DeclareMathOperator*{\argmax}{arg\,max}

\newcommand{\bleu}{{{\sc Bleu}}\xspace}

\newcommand{\eos}{{\it EOS}\xspace}
\newcommand{\sts}{{{\textsc{Seq2Seq}}}\xspace}

\newcommand{\mmiBD}{{MMI-bidi}\xspace}
\newcommand{\mmiLM}{{MMI-antiLM}\xspace}

\newcommand{\mmiBDv}{{$(1-\lambda) \log p(T|S)+\lambda \log p(S|T)$}\xspace}
\newcommand{\mmiLMv}{{$\log p(T|S)-\lambda U(T)$}\xspace}

\title{A Diversity-Promoting Objective Function for Neural Conversation Models}

\author{Jiwei Li\hspace{1pt}$^{{{1}}*}$ \hspace{.4cm} Michel Galley\hspace{1pt}$^{{2}}$  \hspace{.4cm} Chris Brockett\hspace{1pt}$^{2}$ 
\hspace{.4cm} 
Jianfeng Gao\hspace{1pt}$^{2}$ \hspace{.4cm} Bill Dolan\hspace{1pt}$^{2}$
\\[0.3cm]
{$^1$Stanford University, Stanford, CA, USA} \\
{\tt jiweil@stanford.edu} \\
{$^2$Microsoft Research, Redmond, WA, USA} \\
{\tt \{mgalley,chrisbkt,jfgao,billdol\}@microsoft.com}
}

\date{}

\begin{document}
\maketitle

{\let\thefootnote\relax\footnotetext{* The entirety of this work was conducted at Microsoft.}}

\begin{abstract}
Sequence-to-sequence neural network models for generation of conversational responses tend to generate safe, commonplace responses (e.g., \textit{I don't know}) regardless of the input. 
We suggest that the traditional objective function, i.e., the %
likelihood of output (response) given input (message) 
is unsuited to response generation tasks. Instead we propose using Maximum Mutual Information (MMI) as the objective function in neural models.
Experimental results demonstrate that the proposed MMI models produce more diverse, interesting, and appropriate responses, yielding substantive gains in \bleu scores on two conversational datasets and in human evaluations. 
\end{abstract}

\section{Introduction}
\label{sec:intro}
\begin{comment}
\begin{table}
\setlength{\tabcolsep}{4pt}
\center
\small
\begin{tabular}{ll}\hline
\multicolumn{2}{l}{{\bf Input}: What are you doing?}  \\\hline
-0.86~~I don't know.&-1.09~~Get out of here.\\
-1.03~~I don't know!&-1.09~~I'm going home. \\
-1.06~~Nothing.& -1.09~~Oh my god! \\
-1.09~~Get out of the way.&-1.10~~I'm talking to you.\\\hline\hline
\multicolumn{2}{l}{{\bf Input}: what is your name?} \\\hline 
-0.91~~I don't know. & ... \\
-0.92~~I don't know! & -1.55~~My name is Robert.\\
-0.92~~I don't know, sir.&-1.58~~My name is John. \\
-0.97~~Oh, my god!& -1.59~~My name's John. \\\hline\hline
\multicolumn{2}{l}{{\bf Input}: How old are you?} \\\hline 
-0.79~~I don't know. & ... \\
-1.06~~I'm fine.& -1.64~~Twenty-five. \\
-1.17~~I'm all right. & -1.66~~Five.\\
-1.17~~I'm not sure. & -1.71~~Eight. \\\hline
\end{tabular}
\caption{Responses generated by a 4-layer \sts neural model trained on 20 million conversation pairs taken from the OpenSubtitles dataset.
%
%
Decoding is implemented with beam size set to 200. The top examples are the responses with the highest average probability log-likelihoods in the N-best list. Lower-ranked, less-generic responses were manually chosen.} %
\label{sample:mle}
\end{table}
\end{comment}

%
%
%
%
Conversational agents are of growing importance in facilitating smooth interaction between humans and their electronic devices, 
yet conventional dialog systems continue to face major challenges in the form of robustness, scalability and domain adaptation.
Attention has thus turned to learning conversational patterns from data: researchers have begun to explore data-driven generation of conversational responses within the framework of statistical machine translation (SMT), either phrase-based
 \cite{ritter2011data}, or using neural networks to rerank, or directly in the form of sequence-to-sequence (\sts{}) models \cite{Sordoni2015,shang2015neural,vinyals2015neural,wen-EtAl2015,serban2016}.
\sts{} models offer the promise of scalability and language-independence, together with the capacity to implicitly learn semantic and syntactic relations between pairs, and to capture contextual dependencies \cite{Sordoni2015} in a way not possible with conventional SMT approaches \cite{ritter2011data}.

An engaging response generation system should be able to output grammatical, coherent responses that are diverse and interesting.
In practice, however,  neural conversation models tend to generate trivial or non-committal responses, often involving high-frequency phrases along the lines of \textit{I~don't know} or  \textit{I'm OK} \cite{Sordoni2015,vinyals2015neural,serban2016}. 
Table~\ref{sample:mle} illustrates this phenomenon, showing top outputs from \sts{} models. %
All the top-ranked responses are generic. %
Responses that seem more meaningful or specific can also be found in the N-best lists, but rank much lower.
In part at least, this behavior can be ascribed to the relative frequency of generic responses like 
 \textit{I don't know} in conversational datasets, in contrast with the relative sparsity of more contentful alternative responses.\footnote{In our training dataset from the  OpenSubtitles database (OSDb), $0.45\%$ sentences contain the sequence \textit{I don't know}, a high rate considering huge diversity of this dataset.}
It appears that by optimizing for the likelihood of outputs given inputs, neural models assign high probability to ``safe'' responses.
This objective function, common in related tasks such as machine translation, may be unsuited to generation tasks involving intrinsically diverse outputs.
Intuitively, it seems desirable to take into account not only the dependency of responses on messages, but also the inverse, the likelihood that a message will be provided to a given response. %

\begin{table}
\setlength{\tabcolsep}{4pt}
\center
\small
\begin{tabular}{l@{~}l}\hline
\multicolumn{2}{l}{{\bf Input}: What are you doing?}  \\\hline
$-$0.86~~I don't know.&$-$1.09~~Get out of here.\\
$-$1.03~~I don't know!&$-$1.09~~I'm going home. \\
$-$1.06~~Nothing.& $-$1.09~~Oh my god! \\
$-$1.09~~Get out of the way.&$-$1.10~~I'm talking to you.\\\hline\hline
\multicolumn{2}{l}{{\bf Input}: what is your name?} \\\hline 
$-$0.91~~I don't know. & ... \\
$-$0.92~~I don't know! & $-$1.55~~My name is Robert.\\
$-$0.92~~I don't know, sir.&$-$1.58~~My name is John. \\
$-$0.97~~Oh, my god!& $-$1.59~~My name's John. \\\hline\hline
\multicolumn{2}{l}{{\bf Input}: How old are you?} \\\hline 
$-$0.79~~I don't know. & ... \\
$-$1.06~~I'm fine.& $-$1.64~~Twenty-five. \\
$-$1.17~~I'm all right. & $-$1.66~~Five.\\
$-$1.17~~I'm not sure. & $-$1.71~~Eight. \\\hline
\end{tabular}
\caption{Responses generated by a 4-layer \sts neural model trained on 20 million conversation pairs take from the OpenSubtitles dataset.
Decoding is implemented with beam size set to 200. The top examples are the responses with the highest average probability log-likelihoods in the N-best list. Lower-ranked, less-generic responses were manually chosen.} %
\label{sample:mle}
\end{table}

We propose to capture this intuition by using Maximum Mutual Information (MMI), first introduced in speech recognition \cite{bahl1986,brown1987}, as an optimization objective that measures the mutual dependence between inputs and outputs.
Below, we present practical 
strategies for neural generation models that use MMI as an objective function.
We show that use of MMI results in a clear decrease in the proportion of generic response sequences, generating correspondingly more varied and interesting outputs. 

\section{Related work}
\label{sec:related}

The approach we take here is data-driven and end-to-end. This stands in contrast to conventional dialog systems, which typically are template- or heuristic-driven even where there is a statistical component \cite{levin2000stochastic,oh2000stochastic,ratnaparkhi2002trainable,walker2003trainable,pieraccini2009we,young2010hidden,wang2011improving,banchs2012iris,chen2013empirical,ameixa2014luke,nio2014developing}.

We follow a newer line of investigation, originally introduced by Ritter et al. \shortcite{ritter2011data}, which frames response generation as a statistical machine translation (SMT) problem.
Recent progress in SMT stemming from the use of neural language models \cite{sutskever2014sequence,Gao2014,bahdanau2014neural,luong2014addressing} has inspired attempts to extend these neural techniques to response generation.
Sordoni et al. \shortcite{Sordoni2015} improved upon Ritter et al. \shortcite{ritter2011data} by rescoring the output of a phrasal SMT-based conversation system with a \sts model that incorporates prior context. 
Other researchers have subsequently sought to apply direct end-to-end Seq2Seq models \cite{shang2015neural,vinyals2015neural,wen-EtAl2015,YaoEtAl:2015,serban2016}.
These \sts models are Long Short-Term Memory (LSTM) neural networks \cite{hochreiter1997long} that can implicitly capture compositionality and long-span dependencies. \cite{wen-EtAl2015} attempt to learn response templates from crowd-sourced data, whereas we seek to develop methods that can learn conversational patterns from naturally-occurring data. 

%
%
%
%
%
%
%

%
%
%
%
%
%
%

\begin{comment}
Prior work in generation has sought %
%
%
to use Maximal Marginal Relevance (MMR) 
to produce {\it multiple} outputs that are mutually diverse for non-redundant summary sentences \cite{MMR} or nbest lists \cite{Gimpel2013}.
%
Our goal, however, is to produce a {\it single} non-trivial, non-predictable output and our method does not require identifying lexical overlap to foster diversity.\footnote{Augmenting our technique with MMR-based diversity helped increase lexical but not semantic diversity (e.g., {\it I don't know} vs. {\it I haven't a clue}), and with no gain in performance.} 
\end{comment}

Prior work in generation has sought to increase diversity, but with different goals and techniques. Carbonell and Goldstein \shortcite{MMR} and Gimpel et al. \shortcite{Gimpel2013} produce {\it multiple} outputs that are mutually diverse, either non-redundant summary sentences or N-best lists.
Our goal, however, is to produce a {\it single} non-trivial 
output, and our method does not require identifying lexical overlap to foster diversity.\footnote{Augmenting our technique with MMR-based \cite{MMR} diversity helped increase lexical but not semantic diversity (e.g., {\it I don't know} vs. {\it I haven't a clue}), and with no gain in performance.} 

On a somewhat different task, Mao et al. \shortcite[Section 6]{mao2014deep} utilize a mutual information objective in %
image caption retrieval. Below, we focus on the challenge of using MMI in response generation, comparing the performance of MMI models against maximum likelihood. %

\section{Sequence-to-Sequence Models}
\label{sec:seq2seq}
Given a sequence of inputs  $X=\{x_1,x_2,...,x_{N_x}\}$, an LSTM associates each time step with an input gate, a memory gate and an output gate, 
respectively denoted as $i_k$, $f_k$ and $o_k$.
We distinguish $e$ and $h$ where $e_k$ denotes the vector for an individual text unit (for example, a word or sentence) at time step~$k$ while $h_k$ denotes the vector computed by LSTM model at time $k$ by combining $e_k$ and $h_{k-1}$. 
$c_k$ is the cell state vector at time $k$, and
$\sigma$ denotes the sigmoid function. Then, the vector representation $h_k$ for each time step $k$ is given by:
\begin{eqnarray}
i_k=\sigma (W_i\cdot [h_{k-1},e_k])\\
f_k=\sigma (W_f\cdot [h_{k-1},e_k])\\
o_k=\sigma (W_o\cdot [h_{k-1},e_k])\\
l_k=\text{tanh}(W_l\cdot [h_{k-1},e_k])\\
c_k=f_k\cdot c_{k-1}+i_k\cdot l_k\\
h_{k}^s=o_k\cdot \text{tanh}(c_k)
\end{eqnarray}
where $W_i$, $W_f$, $W_o$, $W_l \in \mathbb{R}^{D\times 2D}$.
In \sts generation tasks, each input $X$ is paired with a sequence of outputs to predict: 
$Y=\{y_1,y_2,...,y_{N_y}\}$. 
The LSTM defines a distribution over outputs and sequentially predicts tokens using a softmax function:
\begin{equation*}
\begin{aligned}
p(Y|X)
&=\prod_{k=1}^{N_y}p(y_k|x_1,x_2,...,x_t,y_1,y_2,...,y_{k-1})\\
&=\prod_{k=1}^{N_y}\frac{\exp(f(h_{k-1},e_{y_k}))}{\sum_{y'}\exp(f(h_{k-1},e_{y'}))}
\end{aligned}
\label{equ-lstm}
\end{equation*}
where $f(h_{k-1}, e_{y_k})$ denotes the activation function between $h_{k-1}$ and $e_{y_k}$, where $h_{k-1}$ is the representation output from the LSTM at time \mbox{$k-1$}. 
Each sentence concludes with a special end-of-sentence symbol \eos. 
Commonly, input and output use different LSTMs with separate compositional parameters to capture different compositional patterns. 

During decoding, the algorithm terminates when an \eos token is predicted.
At each time step, either a greedy approach or beam search can be adopted for word prediction.
Greedy search selects the token with the largest conditional probability, the embedding of which is then combined with preceding output to predict the token at the next step.

\section{MMI Models}
\label{sec:models}
\subsection{Notation}
In the response generation task,
let $S$ denote an input message sequence (source) $S=\{s_1,s_2,...,s_{N_s}\}$ where $N_s$ denotes the number of words in $S$.
Let $T$ (target) denote a sequence in response to source sequence $S$, where $T=\{t_1,t_2,...,t_{N_t},$ \eos{}\}, $N_t$ is the length of the response (terminated by an \eos token) and $t$ denotes a word token
that is associated with a $D$ dimensional distinct word embedding $e_t$. $V$~denotes vocabulary size. 

\subsection{MMI Criterion}
The standard objective function for sequence-to-sequence models is the log-likelihood of target $T$ given source $S$, which at test time yields the statistical decision problem:
\begin{equation}
\hat{T}= \argmax_T \big\{\log p(T|S)\big\}
\label{eqseq2seq}
\end{equation}
As discussed in the introduction, we surmise that this formulation leads to generic responses being generated, since it only selects for targets given sources, not the converse. 
To remedy this, we replace it with Maximum Mutual Information (MMI) as the objective function. 
In MMI, parameters are chosen to maximize (pairwise) mutual information
between the source $S$ and the target $T$:
\begin{equation}
\log \frac{p(S,T)}{p(S)p(T)}
\label{eq1}
\end{equation}
This avoids favoring responses that unconditionally enjoy high probability, and instead biases towards those responses that are specific to the given input.
The MMI objective can written as follows:\footnote{Note: $\log\frac{p(S,T)}{p(S)p(T)} = \log\frac{p(T|S)}{p(T)} = \log p(T|S) - \log p(T)$}
\begin{equation*}
\begin{aligned}
\hat{T} = \argmax_{T} \big\{\log p(T|S) - \log p(T)\big\}
\end{aligned}
\end{equation*}
We use a generalization of the MMI objective which introduces a hyperparameter $\lambda$ that controls how much to penalize generic responses:
\begin{equation}
\hat{T}  =\argmax_{T} \big\{\log p(T|S) - \lambda\log p(T)\big\}
\label{eqweightedmmi}
\end{equation}

An alternate formulation of the MMI objective uses Bayes' theorem:
\begin{equation*}
\log p(T)=\log p(T|S)+\log p(S)-\log p(S|T)
\label{eqbayes}
\end{equation*}
which lets us rewrite Equation \ref{eqweightedmmi} as follows:
\begin{equation}
\begin{aligned}
\hat{T}%
&=\argmax_{T} \big\{(1-\lambda)\log p(T|S)\\
&~~~~~~~~~~~~~~~~+\lambda\log p(S|T)-\lambda\log p(S) \big\} \\[0.2cm]
&=\argmax_{T} \big\{(1-\lambda)\log p(T|S)+\lambda\log p(S|T) \big\}
\label{eqbayesexpanded}
\end{aligned}
\end{equation}
This weighted MMI objective function can thus be viewed as representing a tradeoff between sources given targets (i.e., $p(S|T)$) and targets given sources (i.e., $p(T|S)$). 

Although the MMI optimization criterion has been comprehensively studied for other tasks, such as acoustic modeling in speech recognition \cite{huang2001spoken}, adapting MMI to \sts training is empirically nontrivial. 
Moreover, we would like to be able to 
adjust the value $\lambda$ in Equation \ref{eqweightedmmi} without repeatedly training neural network models from scratch, which would otherwise be extremely time-consuming.
Accordingly, we did not train a joint model ($\log p(T|S) - \lambda \log p(T)$), but instead trained maximum likelihood models, and used the MMI criterion only during testing.

\subsection{Practical Considerations}
Responses can be generated either from Equation~\ref{eqweightedmmi},  i.e., $\log p(T|S)-\lambda \log p(T)$ or Equation~\ref{eqbayesexpanded}, i.e., $(1-\lambda) \log p(T|S)+\lambda \log p(S|T)$. 
We will refer to these formulations as \mmiLM and \mmiBD, respectively.
However, these strategies are difficult to apply directly to decoding since they can lead to ungrammatical responses 
(with \mmiLM)
or make decoding intractable 
(with \mmiBD).
In the rest of this section, we will discuss these issues and explain how we resolve them in practice. 

\subsubsection{\mmiLM}
The second term of $\log p(T|S)-\lambda \log p(T)$
functions as an anti-language model.
It penalizes not only high-frequency, generic responses, but also fluent ones and thus can lead to ungrammatical outputs. 
In theory, this issue should not arise when $\lambda$ is less than~1, since ungrammatical sentences should always be more severely penalized by the first term of the equation, i.e., $\log p(T|S)$.
 In practice, however, we found that the model tends to select ungrammatical outputs that escaped being penalized by $p(T|S)$. 
 
\paragraph{Solution}
Again, let $N_t$ be the length of target $T$.
$p(T)$ in Equation~\ref{eqweightedmmi} can be written as:
\begin{equation}
p(T)=\prod_{k=1}^{N_t}p(t_k|t_1,t_2,...,t_{k-1})
\end{equation}
We replace the language model $p(T)$ with $U(T)$, which adapts the standard language model by multiplying by a weight $g(k)$ that is decremented monotonically as the index of the current token $k$ increases:
\begin{equation}
\begin{aligned}
U(T)=\prod_{i=1}^{N_t}p(t_k|t_1,t_2,...,t_{k-1})\cdot g(k)
\end{aligned}
\label{eqmonotonic}
\end{equation}
The underlying intuition here is as follows.
First, neural decoding combines the previously built representation with the word predicted at the current step. 
As decoding proceeds, the influence of the initial input on decoding (i.e., the source sentence representation) diminishes as additional previously-predicted words are encoded in the vector representations.\footnote{Attention models \cite{xu2015show} may offer some promise of addressing this issue.}
In other words, the first words to be predicted significantly determine the remainder of the sentence. 
Penalizing words predicted early on by the language model contributes more to the diversity of the sentence than it does to words predicted later. 
Second, as the influence of the input on decoding declines, 
the influence of the language model comes to dominate. 
We have observed that ungrammatical segments tend to appear in the later parts of the sentences, especially in long sentences. 

We adopt the most straightforward form of $g(k)$
by setting up a threshold ($\gamma$) by penalizing the first $\gamma$ words where\footnote{We experimented with a smooth decay in $g(k)$ rather than a stepwise function, but this did not yield better performance.}
\begin{equation}
\begin{aligned}
g(k)
= \left\{
\begin{aligned}
&1~~~~~\text{if}~k\leq\gamma \\
&0~~~~~\text{if}~k>\gamma
\end{aligned}
\right.
\end{aligned}
\end{equation}
The objective in Equation \ref{eqweightedmmi} 
can thus be rewritten as: 
\begin{equation}
\begin{aligned}
\log p(T|S) - \lambda \log U(T)
\end{aligned}
\label{eqweightedpos}
\end{equation}
where direct decoding is tractable.

\subsubsection{\mmiBD}
Direct decoding from \mmiBDv is intractable, as the second part (i.e., $p(S|T)$) requires completion of 
target generation \textit{before} $p(S|T)$ can be effectively computed.
Due to the enormous search space for target $T$, exploring all possibilities is infeasible. 

For practical reasons, then, we turn to an approximation approach that involves first generating \mbox{N-best} lists given the first part of objective function, i.e., 
standard \sts model $p(T|S)$. 
Then we rerank the \mbox{N-best} lists using the second term of the objective function. 
Since N-best lists produced by \sts models are generally grammatical, the final selected options are likely to be well-formed. 
Model reranking has obvious drawbacks. It results in non-globally-optimal solutions by first emphasizing standard \sts objectives. 
Moreover, it relies heavily on the system's success in generating a sufficiently diverse N-best set, requiring that a long list of N-best lists be generated for each message. 

Nonetheless, these two variants of the MMI criterion work well in practice, significantly improving both interestingness and diversity.

\subsection{Training}

Recent research has shown that deep LSTMs work better than single-layer LSTMs for \sts tasks \cite{vinyals2014grammar,sutskever2014sequence}.
We adopt a deep structure with four LSTM layers for encoding and four LSTM layers for decoding, each of which consists of a different set of parameters. 
Each LSTM layer consists of 1,000 hidden neurons, and the dimensionality of word embeddings is set to 1,000. 
Other training details are given below, broadly aligned with \newcite{sutskever2014sequence}. 
\begin{itemize}[noitemsep,nolistsep]
\item LSTM parameters and embeddings are initialized from a uniform distribution in [$-0.08$, $0.08$].
\item Stochastic gradient decent is implemented using a fixed learning rate of 0.1. 
\item Batch size is set to 256.
\item Gradient clipping is adopted by  scaling gradients when the norm exceeded a threshold of 1. 
\end{itemize}
Our implementation on a single GPU processes at a speed of approximately 600-1200 tokens per second on a Tesla K40. %

The $p(S|T)$ model described in Section 4.3.1 was trained using the same model as that of $p(T|S)$, with messages ($S$) and responses ($T$) interchanged.

\subsection{Decoding}
\subsubsection{\mmiLM}
As described in Section 4.3.1, decoding using \mmiLMv %
 can be readily implemented by predicting tokens at each time-step.
 In addition, we found in our experiments that it is also important to take into account the length of responses in decoding.
 We thus linearly combine the loss function with length penalization, leading to an ultimate score for a given target $T$ as follows:
 \begin{equation}
 Score(T)=p(T|S)-\lambda U(T)+\gamma N_t
 \end{equation}
where $N_t$ denotes the length of the target and $\gamma$ denotes associated weight. We optimize $\gamma$ and $\lambda$ using MERT \cite{mert} on N-best lists of response candidates.
The N-best lists  are generated using the decoder with beam size $B=200$.
We set a maximum length of 20 for generated candidates. 
At each time
step of decoding, we are presented with $B\times B$
candidates. We first add all hypotheses with
an \eos{} token being generated at current time step
to the N-best list. Next we preserve the top $B$ 
unfinished hypotheses and move to next time step.
We therefore maintain beam size of 200 constant
when some hypotheses are completed and taken
down by adding in more unfinished hypotheses.
This will lead the size of final N-best list for each
input much larger than the beam size.

\subsubsection{\mmiBD}
We  generate N-best lists based on $P(T|S)$ and then rerank the list by linearly combining $p(T|S)$, $\lambda p(S|T)$, and $\gamma N_t$. We use MERT to tune the 
weights $\lambda$ and $\gamma$ on the development set.\footnote{As with \mmiLM, we could have used grid search instead of MERT, 
since there are only 3 features and 2 free parameters. In either case, the search attempts to find the best tradeoff between $p(T|S)$ and $p(S|T)$ according to \bleu (which tends to weight the two models relatively equally) and ensures that generated responses are of reasonable length.}

\section{Experiments}
\label{sec:experiments}
\begin{table*}
\center
\small
\begin{tabular}{l|c|c|c|c}
Model                                  &$\#$ of training instances        &\bleu         &{\it distinct-1}      &{\it distinct-2}\\\hline
\sts (baseline)                        & 23M                              & 4.31         & .023               & .107\\
\sts (greedy)                          & 23M                              & 4.51         & .032               & .148\\\hline
\multirow{1}{*}{\mmiLM: \mmiLMv}& \multirow{1}{*}{23M}          &  4.86   & .033              & .175\\
\multirow{1}{*}{\mmiBD: \mmiBDv}&\multirow{1}{*}{23M}& {\bf 5.22} & .051               & .270\\ \hline
SMT \cite{ritter2011data}              & 50M                              & 3.60         & .098               & .351\\
SMT+neural reranking \cite{Sordoni2015}& 50M                              & 4.44         & {\bf .101}         & {\bf .358}\\\hline
\end{tabular}
\caption{Performance on the Twitter dataset of 4-layer \sts models and MMI models. {\it distinct-1} and {\it distinct-2} are respectively the number of distinct unigrams and bigrams divided by total number of generated words.}
\label{res:twitter}
\end{table*}

\subsection{Datasets}
\paragraph{Twitter Conversation Triple Dataset} 
We used an extension of the dataset described in Sordoni et al.
\shortcite{Sordoni2015}, which consists of 23 million conversational snippets randomly selected from a collection of 129M context-message-response triples extracted from the Twitter Firehose over the 3-month period from June through August 2012.
For the purposes of our experiments, we limited context to the turn in the conversation immediately preceding the message. In our LSTM models, we used a simple input model in which contexts and messages are concatenated to form the source input. 

For tuning and evaluation, we used the development dataset (2118 conversations) and the test dataset (2114 examples), augmented using information retrieval methods to create a multi-reference set, as described by Sordoni et al. \shortcite{Sordoni2015}. 
The selection criteria for these two datasets included a component of relevance/interestingness, with the result that dull responses will tend to be penalized in evaluation.

\paragraph{OpenSubtitles dataset} In addition to unscripted Twitter conversations, we also used the OpenSubtitles (OSDb) dataset \cite{tiedemann2009news}, a large, noisy, open-domain dataset containing roughly 60M-70M scripted lines spoken by movie characters. 
This dataset does not specify which character speaks each subtitle line, which prevents us from inferring speaker turns. 
Following Vinyals et al. (2015), we make the simplifying assumption that each line of subtitle constitutes a full speaker turn. Our models are trained to predict the current turn given the preceding ones based on the assumption that consecutive turns belong to the same conversation.
This introduces a degree of noise, since consecutive lines may not appear in the same conversation or scene, and may not even be spoken by the same character.

This limitation potentially renders the OSDb dataset unreliable for evaluation purposes.  
For evaluation purposes, we therefore used data from the Internet Movie Script Database (IMSDB),\footnote{IMSDB (\url{http://www.imsdb.com/}) is a relatively small database of around 0.4 million sentences and thus not suitable for open domain dialogue training.} which explicitly identifies which character speaks each line of the script. 
This allowed us to identify consecutive message-response pairs spoken by different characters. %
We randomly selected two subsets as development and test datasets, each containing 2k pairs, with source and target length restricted to the range of [6,18]. 

\begin{table}
\center
\small
\begin{tabular}{c|c|c|c}
Model&\bleu&{\it distinct-1}&{\it distinct-2}\\\hline
\sts&1.28&0.0056&0.0136\\\hline
\mmiLM&1.74 & 0.0184&0.066\\
&(+35.9$\%$)&(+228$\%$)&(407$\%$)\\\hline
\mmiBD&1.44&0.0103&0.0303\\
&(+28.2$\%$)&(+83.9$\%$)&(+122$\%$)\\\hline
\end{tabular}
\caption{Performance of the \sts baseline and two MMI models on the OpenSubtitles dataset.}
\label{res:open}
\end{table}

\subsection{Evaluation}
For parameter tuning and final evaluation,
we used \bleu \cite{Papineni2002BLEU}, which was shown to correlate reasonably well with human judgment on the response generation task \cite{galley2015deltableu}.
In the case of the Twitter models, we used multi-reference \bleu. 
As the IMSDB data is too limited to support extraction of multiple references, only single reference \bleu was used in training and evaluating the OSDb models.

We did not follow Vinyals and Le \shortcite{vinyals2015neural} in using perplexity as evaluation metric. 
Perplexity is unlikely to be a useful metric in our scenario, since our proposed model is designed to steer away from the standard \sts model in order to diversify the outputs.   
We report degree of diversity by calculating the number of distinct unigrams and bigrams in generated responses. The value is scaled by total number of generated tokens to avoid favoring long sentences (shown as {\it distinct-1} and {\it distinct-2} in Tables \ref{res:twitter} and \ref{res:open}). 

\subsection{Results}

\paragraph{Twitter Dataset} We first report performance on Twitter datasets in Table~\ref{res:twitter}, along with results
for different models (i.e., {\it Machine Translation} and {\it MT+neural reranking})
reprinted from Sordoni et al. \shortcite{Sordoni2015} on the same dataset. The baseline is the \sts model with its standard likelihood objective and a beam size of 200. We compare this baseline against greedy-search \sts \cite{vinyals2015neural}, which can help achieve higher diversity by increasing search errors.\footnote{Another method would have been to sample from the $p(T|S)$ distribution to increase diversity. While these methods have merits, we think we ought to find a proper objective and optimize it exactly, rather than cope with an inadequate one and add noise to it.}
 
{\it Machine Translation} is the phrase-based MT system described in  
\cite{ritter2011data}. MT features include commonly used ones in Moses \cite{koehn2007moses}, e.g., 
forward and backward maximum
likelihood ``translation'' probabilities, word and
phrase penalties, linear distortion, etc. For more details, refer to Sordoni et al. \shortcite{Sordoni2015}. 

{\it MT+neural reranking} is the phrase-based MT system, reranked using neural models.  
N-best lists are first generated from the MT system.
Recurrent neural models generate scores for N-best list candidates given the input messages.
These generated scores are re-incorporated to rerank all the candidates. 
Additional features to score [1-4]-gram matches between context and response and between message and context (context and message match CMM features) are also employed, as in Sordoni et al. \shortcite{Sordoni2015}.  

{\it MT+neural reranking} achieves a \bleu score of 4.44, which to the best of our knowledge represents the previous state-of-the-art performance on this Twitter dataset.  Note that  
{\it Machine Translation}  and {\it MT+neural reranking}  are trained on a much larger dataset of roughly 50 million examples.
A~significant performance boost is observed from \mmiBD over baseline \sts, both in terms of \bleu score and diversity.  

The beam size of 200 used in our main experiments is quite conservative, and \bleu scores only slightly degrade when
reducing beam size to 20. For \mmiBD, \bleu scores for beam sizes of 200, 50, 20 are respectively 5.90, 5.86, 5.76.
A beam size  of 20 still produces relatively large N-best lists (173 elements on average) with responses of varying lengths, which offer enough diversity for the $p(S|T)$ model to have a significant effect.

\paragraph{OpenSubtitles Dataset}
All models achieve significantly lower \bleu scores on this dataset than on the Twitter dataset, primarily because the IMSDB data provides only single references for evaluation. 
We note, however, that baseline \sts models yield lower levels of unigram diversity ({\it distinct-1}) on the OpenSubtitles dataset than on the Twitter data (0.0056 vs 0.017), which suggests that other factors may be in play. 
It is likely that movie dialogs are much more concise and information-rich than typical conversations on Twitter, making it harder to match gold-standard responses and causing the learned models to strongly favor safe, conservative responses.
 
Table \ref{res:open} shows that the \mmiLM model
yields a significant performance boost, with a \bleu score increase of up
to $36\%$ and a more than $200\%$ jump in unigram diversity. 
Our interpretation of this huge performance improvement is that the
diversity and complexity of input messages lead 
standard \sts models to generate very conservative responses,\footnote{The strings 
\textit{I don't know}, \textit{I don't know what you are talking about},
\textit{I don't think that is a good idea}, and
\textit{Oh my god} constitute
$32\%$ percent of all generated responses.} which fail to match the more interesting reference strings typical of this dataset.
This interpretation is also supported by the fact that the
\mmiBD model does not produce as significant a performance boost as \mmiLM. 
In the case of \mmiBD, N-best lists generated using standard \sts models remain conservative and uninteresting, attenuating the impact of later reranking.  
An important potential limitation of \mmiBD model is thus that its performance hinges on the initial generation of a highly diverse, informative N-best list. 

\begin{table*}
\center
\footnotesize
\begin{tabular}{p{5cm}p{5cm}p{5cm} }\hline
message& {\sts} &MMI $P(T|S)-\lambda P(T)$\\\hline
I did not get the report from an mi-6 agent.&I don't think that's a good idea. &
You did the right thing, did you?
 \\\hline
You haven't been given an assignment in this case. &
I don't know what you are talking about.&
I've been looking all over for you. \\\hline
I'm losing my grip.&
I don't know what you are talking about.&
I'm the only one in the world. \\\hline
I am ready to help. &
Come on, come on &
I have something we need to talk about.\\\hline
You programmed me to gather intelligence. That's all I've ever done. &
You know that, don't you? &
You do have fun, don't you? \\\hline
I mean, we'd have to talk to him.&
I mean, I don't know. &
I mean, he's a good guy.\\\hline
This has nothing to do with him.&
You don't know anything about him, do you? &
I know him as much.\\\hline
How come you never say it? & I don't know & Because I don't want to hurt you\\\hline
He is the one behind this scheme to crash x-net. & I don't know what you are talking about &Yes, he is, isn't he?\\\hline
How much time do you have here?&I don't know.& Not long enough. Sorry, sir.\\\hline
\end{tabular}
\caption{Responses from the \sts baseline and \mmiLM models on the OpenSubtitles dataset.}
\label{out:model1}
\end{table*}

\begin{table*}[!ht]
\center
\footnotesize
\begin{tabular}{p{5cm}p{5cm}p{5cm} }\hline
message & {\sts} & \mmiBD\\\hline
Ohh I've never seen that! How long does it take you guys to learn the drill?& I don't even know.&Like 2 weeks ago!!\\\hline
You'll know when we sign Heskey in 2 weeks&I'll see what I can do&
I sure hope so..\\\hline
I'm waiting!! I have even got biscuits waiting for you??&
I have no idea what you're talking about&
I'll send you a pic when I get home xx \\\hline
My parents actually had touch too much played at their wedding.&
I have no idea what you're talking about&
How old are they?? \\\hline
Say anything you want... I know you love me and I love you.&I love you so much.& I love you too, baby. \\\hline
I am off all week next week & What time you going? & What time you going?\\\hline
How are you doing? & I'm good, thanks.&I'm good, you?\\\hline
\end{tabular}
\caption{Responses from the \sts baseline and \mmiBD models on the Twitter dataset.}
\label{out:model2}
\end{table*}

\begin{table}[tbp]
\small
\centering
\begin{tabular}{@{}lcc@{}}
\toprule
{\bf Comparator} & {\bf Gain} & {\bf 95\% CI}\\
\midrule
SMT \cite{ritter2011data} & 0.29 & [0.25, 0.32] \\
SMT+neural reranking & 0.28 & [0.25, 0.32] \\
\midrule
\sts (baseline) & 0.11 & [0.07, 0.14] \\
\sts (greedy) & 0.08 & [0.04, 0.11] \\
\bottomrule
\end{tabular}
\caption{\mmiBD gains over comparator systems, based on pairwise human judgments.}
\label{tab:humanscores}
\end{table}

\begin{table}
\setlength{\tabcolsep}{4pt}
\center
\footnotesize
\begin{tabular}{ll}\hline
\multicolumn{2}{l}{{\bf Input}: What are you doing?} \\\hline
1. I've been looking for you.&4. I told you to shut up.\\
2. I want to talk to you. &5. Get out of here.\\
3. Just making sure you're OK.&6. I'm looking for a doctor. \\\hline
\multicolumn{2}{l}{{\bf Input}: What is your name? }\\\hline
1. Blue! & 4. Daniel. \\
2. Peter. &5. My name is John. \\
3. Tyler. &6. My name is Robert. \\\hline
\multicolumn{2}{l}{{\bf Input}: How old are you?} \\\hline
1. Twenty-eight. & 4. Five.\\
2. Twenty-four. & 5. 15.\\
3. Long.& 6. Eight.\\\hline
\end{tabular}
\caption{Examples generated by the \mmiLM model on the OpenSubtitles dataset.} 
\label{sample:mmi}
\end{table}

\paragraph{Qualitative Evaluation}

We employed crowdsourced judges to provide evaluations for a random sample of 1000 items in the Twitter test dataset. 
Table \ref{tab:humanscores} shows the results of human evaluations between paired systems. 
Each output pair was ranked by 5 judges, who were asked to decide which of the two outputs was better. 
They were instructed to prefer outputs that were more specific (relevant) to the message and preceding context, as opposed to those that were more generic. 
Ties were permitted. 
Identical strings were algorithmically assigned the same score. 
The mean of differences between outputs is shown as the gain for \mmiBD over the competing system.
At a significance level of $\alpha = 0.05$, we find that \mmiBD outperforms both baseline and greedy \sts systems, as well as the weaker SMT and SMT+RNN baselines. 
\mmiBD outperforms SMT in human evaluations \textit{despite} the greater lexical diversity of MT output. %

Separately, judges were also asked to rate overall quality of \mmiBD output over the same 1000-item sample in isolation, each output being evaluated by 7 judges in context using a 5-point scale. 
The mean rating was 3.84 (median: 3.85, 1st Qu: 3.57, 3rd Qu: 4.14), suggesting that overall \mmiBD output does appear reasonably acceptable to human judges.\footnote{In the human evaluations, we asked the annotators to prefer responses that were more specific to the context only when doing the pairwise evaluations of systems. The absolute evaluation was conducted separately (on different days) on the best system, and annotators were asked to evaluate the overall quality of the response, specifically \textit{Provide your impression of overall quality of the response in this particular conversation.}}

Table~\ref{sample:mmi} presents the N-best candidates generated using the \mmiBD model for the inputs of Table~\ref{sample:mle}. We see that MMI generates significantly more interesting outputs than \sts. 
 
In Tables~\ref{out:model1} and \ref{out:model2}, we present responses generated by different models.
All examples were randomly sampled (without cherry picking).
We see that the baseline \sts model tends to generate reasonable responses to simple messages such as 
\textit{How are you doing?} or \textit{I love you}. 
As the complexity of the message increases, however, the outputs switch to more conservative, duller forms, such as 
\textit{I don't know} or \textit{I don't know what you are talking about}.
An occasional answer of this kind might go unnoticed in a natural conversation, but a dialog agent that \textit{always} produces such responses risks being perceived as uncooperative. 
\mmiBD models, on the other hand, produce far more diverse and interesting responses.

\section{Conclusions}
\label{sec:conclusion}

We investigated an issue encountered when applying \sts models to conversational response generation. 
These models tend to generate safe, commonplace responses (e.g., \textit{I don't know}) regardless of the input. 
Our analysis suggests that the issue is at least in part attributable to the use of 
unidirectional likelihood of output (responses) given input (messages).
To remedy this, we have proposed using Maximum Mutual Information (MMI) as the objective function.
Our results demonstrate that the proposed MMI models produce more diverse and interesting responses, while improving quality as measured by \bleu and human evaluation. 

To the best of our knowledge, this paper represents the first work to address the issue of output diversity in the neural generation framework. 
We have focused on the algorithmic dimensions of the problem.  
Unquestionably numerous other factors such as grounding, persona (of both user and agent), and intent also play a role 
in generating diverse, conversationally interesting outputs. 
These must be left for future investigation.   
Since the challenge of producing interesting outputs also arises in other neural generation tasks, including image-description generation, question answering, and potentially any task where mutual correspondences must be modeled, the implications of this work extend well beyond conversational response generation.

\section*{Acknowledgments}

We thank the anonymous reviewers, as well as 
Dan Jurafsky, 
Alan Ritter, 
Stephanie Lukin, 
George Spithourakis, 
Alessandro Sordoni,
Chris Quirk, 
Meg Mitchell,
Jacob Devlin, 
Oriol Vinyals, and
Dhruv Batra 
for their comments and suggestions.

\bibliographystyle{naaclhlt2016}
\bibliography{tacl-llr-clean}

\begin{thebibliography}{}

\bibitem[\protect\citename{Ameixa \bgroup et al.\egroup }2014]{ameixa2014luke}
David Ameixa, Luisa Coheur, Pedro Fialho, and Paulo Quaresma.
\newblock 2014.
\newblock Luke, {I} am your father: dealing with out-of-domain requests by
  using movies subtitles.
\newblock In {\em Intelligent Virtual Agents}, pages 13--21. Springer.

\bibitem[\protect\citename{Bahdanau \bgroup et al.\egroup
  }2015]{bahdanau2014neural}
Dzmitry Bahdanau, Kyunghyun Cho, and Yoshua Bengio.
\newblock 2015.
\newblock Neural machine translation by jointly learning to align and
  translate.
\newblock In {\em Proc. of the International Conference on Learning
  Representations (ICLR)}.

\bibitem[\protect\citename{Bahl \bgroup et al.\egroup }1986]{bahl1986}
L.~Bahl, P.~Brown, P.~de~Souza, and R.~Mercer.
\newblock 1986.
\newblock {Maximum mutual information estimation of hidden Markov model
  parameters for speech recognition}.
\newblock {\em Acoustics, Speech, and Signal Processing, IEEE International
  Conference on ICASSP '86.}, pages 49--52.

\bibitem[\protect\citename{Banchs and Li}2012]{banchs2012iris}
Rafael~E Banchs and Haizhou Li.
\newblock 2012.
\newblock {IRIS}: a chat-oriented dialogue system based on the vector space
  model.
\newblock In {\em Proc. of the ACL 2012 System Demonstrations}, pages 37--42.

\bibitem[\protect\citename{Brown}1987]{brown1987}
Peter~F. Brown.
\newblock 1987.
\newblock {\em The Acoustic-modeling Problem in Automatic Speech Recognition}.
\newblock {Ph.D.} thesis, Carnegie Mellon University.

\bibitem[\protect\citename{Carbonell and Goldstein}1998]{MMR}
Jaime Carbonell and Jade Goldstein.
\newblock 1998.
\newblock The use of mmr, diversity-based reranking for reordering documents
  and producing summaries.
\newblock In {\em In Research and Development in Information Retrieval}, pages
  335--336.

\bibitem[\protect\citename{Chen \bgroup et al.\egroup }2013]{chen2013empirical}
Yun-Nung Chen, Wei~Yu Wang, and Alexander Rudnicky.
\newblock 2013.
\newblock An empirical investigation of sparse log-linear models for improved
  dialogue act classification.
\newblock In {\em Proc. of ICASSP}, pages 8317--8321.

\bibitem[\protect\citename{Galley \bgroup et al.\egroup
  }2015]{galley2015deltableu}
Michel Galley, Chris Brockett, Alessandro Sordoni, Yangfeng Ji, Michael Auli,
  Chris Quirk, Margaret Mitchell, Jianfeng Gao, and Bill Dolan.
\newblock 2015.
\newblock \dbleu: A discriminative metric for generation tasks with
  intrinsically diverse targets.
\newblock In {\em Proc. of ACL-IJCNLP}, pages 445--450, Beijing, China, July.

\bibitem[\protect\citename{Gao \bgroup et al.\egroup }2014]{Gao2014}
Jianfeng Gao, Xiaodong He, Wen-tau Yih, and Li~Deng.
\newblock 2014.
\newblock Learning continuous phrase representations for translation modeling.
\newblock In {\em Proc. of ACL}, pages 699--709.

\bibitem[\protect\citename{Gimpel \bgroup et al.\egroup }2013]{Gimpel2013}
Kevin Gimpel, Dhruv Batra, Chris Dyer, and Gregory Shakhnarovich.
\newblock 2013.
\newblock A systematic exploration of diversity in machine translation.
\newblock In {\em Proceedings of the 2013 Conference on Empirical Methods in
  Natural Language Processing}, pages 1100--1111.

\bibitem[\protect\citename{Hochreiter and Schmidhuber}1997]{hochreiter1997long}
Sepp Hochreiter and J{\"u}rgen Schmidhuber.
\newblock 1997.
\newblock Long short-term memory.
\newblock {\em Neural computation}, 9(8):1735--1780.

\bibitem[\protect\citename{Huang \bgroup et al.\egroup }2001]{huang2001spoken}
Xuedong Huang, Alex Acero, Hsiao-Wuen Hon, and Raj Foreword By-Reddy.
\newblock 2001.
\newblock {\em Spoken language processing: A guide to theory, algorithm, and
  system development}.
\newblock Prentice Hall.

\bibitem[\protect\citename{Koehn \bgroup et al.\egroup }2007]{koehn2007moses}
Philipp Koehn, Hieu Hoang, Alexandra Birch, Chris Callison-Burch, Marcello
  Federico, Nicola Bertoldi, Brooke Cowan, Wade Shen, Christine Moran, Richard
  Zens, Chris Dyer, Ondrej Bojar, Alexandra Constantin, and Evan Herbst.
\newblock 2007.
\newblock Moses: Open source toolkit for statistical machine translation.
\newblock In {\em Proc. of the 45th Annual Meeting of the Association for
  Computational Linguistics}, pages 177--180, Prague, Czech Republic, June.
  Association for Computational Linguistics.

\bibitem[\protect\citename{Levin \bgroup et al.\egroup
  }2000]{levin2000stochastic}
Esther Levin, Roberto Pieraccini, and Wieland Eckert.
\newblock 2000.
\newblock A stochastic model of human-machine interaction for learning dialog
  strategies.
\newblock {\em IEEE Transactions on Speech and Audio Processing}, 8(1):11--23.

\bibitem[\protect\citename{Luong \bgroup et al.\egroup
  }2015]{luong2014addressing}
Thang Luong, Ilya Sutskever, Quoc Le, Oriol Vinyals, and Wojciech Zaremba.
\newblock 2015.
\newblock Addressing the rare word problem in neural machine translation.
\newblock In {\em Proc. of ACL-IJCNLP}, pages 11--19, Beijing, China.

\bibitem[\protect\citename{Mao \bgroup et al.\egroup }2015]{mao2014deep}
Junhua Mao, Wei Xu, Yi~Yang, Jiang Wang, Zhiheng Huang, and Alan Yuille.
\newblock 2015.
\newblock Deep captioning with multimodal recurrent neural networks (m-{RNN}).
\newblock {\em ICLR}.

\bibitem[\protect\citename{Nio \bgroup et al.\egroup }2014]{nio2014developing}
Lasguido Nio, Sakriani Sakti, Graham Neubig, Tomoki Toda, Mirna Adriani, and
  Satoshi Nakamura.
\newblock 2014.
\newblock Developing non-goal dialog system based on examples of drama
  television.
\newblock In {\em Natural Interaction with Robots, Knowbots and Smartphones},
  pages 355--361. Springer.

\bibitem[\protect\citename{Och}2003]{mert}
Franz~Josef Och.
\newblock 2003.
\newblock Minimum error rate training in statistical machine translation.
\newblock In {\em Proceedings of the 41st Annual Meeting of the Association for
  Computational Linguistics}, pages 160--167, Sapporo, Japan, July. Association
  for Computational Linguistics.

\bibitem[\protect\citename{Oh and Rudnicky}2000]{oh2000stochastic}
Alice~H Oh and Alexander~I Rudnicky.
\newblock 2000.
\newblock Stochastic language generation for spoken dialogue systems.
\newblock In {\em Proc. of the 2000 ANLP/NAACL Workshop on Conversational
  systems-Volume 3}, pages 27--32.

\bibitem[\protect\citename{Papineni \bgroup et al.\egroup
  }2002]{Papineni2002BLEU}
Kishore Papineni, Salim Roukos, Todd Ward, and Wei-Jing Zhu.
\newblock 2002.
\newblock {\sc Bleu}: a method for automatic evaluation of machine translation.
\newblock In {\em Proc. of ACL}.

\bibitem[\protect\citename{Pieraccini \bgroup et al.\egroup
  }2009]{pieraccini2009we}
Roberto Pieraccini, David Suendermann, Krishna Dayanidhi, and Jackson Liscombe.
\newblock 2009.
\newblock Are we there yet? research in commercial spoken dialog systems.
\newblock In {\em Text, Speech and Dialogue}, pages 3--13. Springer.

\bibitem[\protect\citename{Ratnaparkhi}2002]{ratnaparkhi2002trainable}
Adwait Ratnaparkhi.
\newblock 2002.
\newblock Trainable approaches to surface natural language generation and their
  application to conversational dialog systems.
\newblock {\em Computer Speech \& Language}, 16(3):435--455.

\bibitem[\protect\citename{Ritter \bgroup et al.\egroup }2011]{ritter2011data}
Alan Ritter, Colin Cherry, and William Dolan.
\newblock 2011.
\newblock Data-driven response generation in social media.
\newblock In {\em Proc. of EMNLP}, pages 583--593.

\bibitem[\protect\citename{Serban \bgroup et al.\egroup }2016]{serban2016}
Iulian~V Serban, Alessandro Sordoni, Yoshua Bengio, Aaron Courville, and Joelle
  Pineau.
\newblock 2016.
\newblock Building end-to-end dialogue systems using generative hierarchical
  neural network models.
\newblock In {\em Proc. of AAAI}, February.

\bibitem[\protect\citename{Shang \bgroup et al.\egroup }2015]{shang2015neural}
Lifeng Shang, Zhengdong Lu, and Hang Li.
\newblock 2015.
\newblock Neural responding machine for short-text conversation.
\newblock In {\em Proc. of ACL-IJCNLP}, pages 1577--1586.

\bibitem[\protect\citename{Sordoni \bgroup et al.\egroup }2015]{Sordoni2015}
Alessandro Sordoni, Michel Galley, Michael Auli, Chris Brockett, Yangfeng Ji,
  Meg Mitchell, Jian-Yun Nie, Jianfeng Gao, and Bill Dolan.
\newblock 2015.
\newblock A neural network approach to context-sensitive generation of
  conversational responses.
\newblock In {\em Proc. of NAACL-HLT}, May--June.

\bibitem[\protect\citename{Sutskever \bgroup et al.\egroup
  }2014]{sutskever2014sequence}
Ilya Sutskever, Oriol Vinyals, and Quoc Le.
\newblock 2014.
\newblock Sequence to sequence learning with neural networks.
\newblock In {\em Proc. of NIPS}, pages 3104--3112.

\bibitem[\protect\citename{Tiedemann}2009]{tiedemann2009news}
J{\"o}rg Tiedemann.
\newblock 2009.
\newblock News from {OPUS} -- a collection of multilingual parallel corpora
  with tools and interfaces.
\newblock In {\em Recent advances in natural language processing}, volume~5,
  pages 237--248.

\bibitem[\protect\citename{Vinyals and Le}2015]{vinyals2015neural}
Oriol Vinyals and Quoc Le.
\newblock 2015.
\newblock A neural conversational model.
\newblock In {\em Proc. of ICML Deep Learning Workshop}.

\bibitem[\protect\citename{Vinyals \bgroup et al.\egroup
  }2015]{vinyals2014grammar}
Oriol Vinyals, Lukasz Kaiser, Terry Koo, Slav Petrov, Ilya Sutskever, and
  Geoffrey Hinton.
\newblock 2015.
\newblock Grammar as a foreign language.
\newblock In {\em Proc. of NIPS}.

\bibitem[\protect\citename{Walker \bgroup et al.\egroup
  }2003]{walker2003trainable}
Marilyn~A Walker, Rashmi Prasad, and Amanda Stent.
\newblock 2003.
\newblock A trainable generator for recommendations in multimodal dialog.
\newblock In {\em INTERSPEECH}.

\bibitem[\protect\citename{Wang \bgroup et al.\egroup }2011]{wang2011improving}
William~Yang Wang, Ron Artstein, Anton Leuski, and David Traum.
\newblock 2011.
\newblock Improving spoken dialogue understanding using phonetic mixture
  models.
\newblock In {\em FLAIRS}.

\bibitem[\protect\citename{Wen \bgroup et al.\egroup }2015]{wen-EtAl2015}
Tsung-Hsien Wen, Milica Gasic, Nikola Mrk\v{s}i\'{c}, Pei-Hao Su, David
  Vandyke, and Steve Young.
\newblock 2015.
\newblock Semantically conditioned {LSTM}-based natural language generation for
  spoken dialogue systems.
\newblock In {\em Proc. of EMNLP}, pages 1711--1721, Lisbon, Portugal,
  September.

\bibitem[\protect\citename{Xu \bgroup et al.\egroup }2015]{xu2015show}
Kelvin Xu, Jimmy Ba, Ryan Kiros, Kyunghyun Cho, Aaron Courville, Ruslan
  Salakhudinov, Rich Zemel, and Yoshua Bengio.
\newblock 2015.
\newblock Show, attend and tell: Neural image caption generation with visual
  attention.
\newblock In {\em Proc. of ICML}, pages 2048--2057. JMLR Workshop and
  Conference Proceedings.

\bibitem[\protect\citename{Yao \bgroup et al.\egroup }2015]{YaoEtAl:2015}
Kaisheng Yao, Geoffrey Zweig, and Baolin Peng.
\newblock 2015.
\newblock Attention with intention for a neural network conversation model.
\newblock In {\em NIPS workshop on Machine Learning for Spoken Language
  Understanding and Interaction}.

\bibitem[\protect\citename{Young \bgroup et al.\egroup }2010]{young2010hidden}
Steve Young, Milica Ga{\v{s}}i{\'c}, Simon Keizer, Fran{\c{c}}ois Mairesse,
  Jost Schatzmann, Blaise Thomson, and Kai Yu.
\newblock 2010.
\newblock The hidden information state model: A practical framework for
  {POMDP}-based spoken dialogue management.
\newblock {\em Computer Speech \& Language}, 24(2):150--174.

\end{thebibliography}

\end{document}